\documentclass[conference]{IEEEtran}
\IEEEoverridecommandlockouts

\usepackage{cite}
\usepackage{amsmath,amssymb,amsfonts}
\usepackage{algorithmic}
\usepackage{graphicx}
\usepackage{textcomp}
\usepackage{xcolor}

\usepackage{booktabs}       
\usepackage{nicefrac}       
\usepackage{microtype}      
\usepackage{amsthm}
\usepackage{multirow}
\usepackage{colortbl}

\usepackage{tabularx}
\usepackage{tikz}
\usepackage{commath}
\usepackage{lipsum}

\def\BibTeX{{\rm B\kern-.05em{\sc i\kern-.025em b}\kern-.08em
    T\kern-.1667em\lower.7ex\hbox{E}\kern-.125emX}}
\begin{document}

\definecolor{lightpastelpurple}{rgb}{0.69, 0.61, 0.85}
\definecolor{gray}{rgb}{0.75, 0.75, 0.75}
\colorlet{LightGreen}{lightpastelpurple!40}
\colorlet{Gray}{gray!40}

\definecolor{maroon}{cmyk}{0,0.87,0.68,0.32}
\newcommand{\gray}{\rowcolor[gray]{.90}}

\title{Representation Learning with Semantic-aware Instance and Sparse Token Alignments
}

\author{
\IEEEauthorblockN{Phuoc-Nguyen Bui, Toan Duc Nguyen, Junghyun Bum, Duc-Tai Le, Hyunseung Choo}
\IEEEauthorblockA{\textit{Sungkyunkwan University} \\
Suwon, Korea
}
}

\maketitle

\begin{abstract}
Medical contrastive vision-language pre-training (VLP) has demonstrated significant potential in improving performance on downstream tasks. Traditional approaches typically employ contrastive learning, treating paired image-report samples as positives and unpaired ones as negatives. However, in medical datasets, there can be substantial similarities between images or reports from different patients. Rigidly treating all unpaired samples as negatives, can disrupt the underlying semantic structure and negatively impact the quality of the learned representations. In this paper, we propose a multi-level alignment framework, Representation Learning with \textbf{S}emantic-aware \textbf{I}nstance and \textbf{S}parse \textbf{T}oken \textbf{A}lignments (\textbf{SISTA}) by exploiting the semantic correspondence between medical image and radiology reports at two levels, i.e., image-report and patch-word levels. Specifically, we improve the conventional contrastive learning by incorporating inter-report similarity to eliminate the false negatives and introduce a method to effectively align image patches with relevant word tokens. Experimental results demonstrate the effectiveness of the proposed framework in improving transfer performance across different datasets on three downstream tasks: image classification, image segmentation, and object detection. Notably, our framework achieves significant improvements in fine-grained tasks even with limited labeled data. Codes and pre-trained models will be made available.
\end{abstract}


\section{Introduction}

Vision-language pre-training (VLP) has emerged as a powerful method for models to learn robust, transferable representations from large image-text datasets. Visual representation learning in this context is generally guided by three principal objectives. Contrastive objectives encourage vision-language models (VLMs) to learn discriminative representations by minimizing the distance between paired image-text embeddings and maximizing separation from unpaired samples in the feature space \cite{chauhan2020joint, radford2021learning, yang2022unified, zhang2022contrastive}. Generative objectives focus on modeling semantic structures by training networks to synthesize data, encompassing tasks such as image reconstruction \cite{he2022masked, li2023general, wei2022masked}, language generation \cite{yu2022coca}, or cross-modal generation \cite{singh2022flava, zhou2023advancing, bui2024visual}. Finally, alignment objectives seek to refine image-text correspondence through either global image-text alignment \cite{dou2022coarse, bao2021vlmo} or localized region-word alignment within the embedding space \cite{yao2022detclip, li2022grounded}, enhancing the model’s capacity to capture fine-grained semantic relationships across modalities.

In the medical domain, where data labeling is both resource-intensive and costly, VLP models have demonstrated potential in improving data efficiency, enabling zero-shot recognition, and facilitating transfer learning. By aligning medical images (e.g., X-rays, CT scans) with their corresponding radiology reports, VLP models enhance performance across downstream tasks such as disease classification, segmentation, and image-based retrieval. Several VLP methods \cite{wang2022medclip, wang2022multi, zhou2022generalized, liu2023improving, lai2024carzero, park2024self, wan2024med} have been specifically developed to tackle the inherent complexities of integrating multimodal data in clinical applications. While these approaches have shown promising results, the conventional contrastive learning paradigms they employ may be suboptimal for medical data due to the unique characteristics of this domain. Medical image-report pairs often exhibit high similarity across cases, i.e. several images may have similar report. Conventional contrastive learning frameworks \cite{wang2022multi, li2024mlip} that treat all unpaired samples as negatives can lead to what we refer to as false negatives—instances where semantically similar images or reports are mistakenly considered negatives simply because they originate from different patient cases. This rigid handling of unpaired samples distorts the semantic structure within the learned representation space, ultimately impairing the model's generalization capability.

To address the limitations of conventional contrastive learning in medical vision-language tasks, we propose a novel multi-level alignment framework, \textbf{SISTA}, which captures both instance-level and token-level correspondences between medical images and radiology reports. Our framework is composed of two core modules: the Semantics-aware Instance-level Alignment (SIA) module, which utilizes inter-report similarity to mitigate the impact of false negatives, and the Sparse Token-level Alignment (STA) module, which learns sparse, selective groupings of image patches to align with corresponding words in radiology reports. Additionally, the SIA module can be applied in a self-supervised manner within each modality to further enhance representation learning. The integration of these modules enables the model to learn robust and transferable representations, particularly effective for fine-grained tasks such as segmentation and detection. Comprehensive experiments show that our framework outperforms prior methods across key downstream tasks, establishing a new benchmark for medical vision-language pre-training. In summary, the contributions of this paper are as follows:

\begin{itemize}
    \item We propose the Semantics-aware Instance-level Alignment module, which enhances conventional contrastive learning by using inter-report relationships to address false negatives. 
    \item We introduce the Sparse Token-level Alignment module, which employs sparse cross-attention to identify and group the most relevant image patches for each word in the report, capturing fine-grained visual-textual correspondences critical for segmentation and detection tasks.
    \item We develop a multi-level alignment framework, namely SISTA, that combines cross-modal and intra-modal SIA’s instance-level alignment with STA’s token-level alignment, enabling the model to learn both global and localized cross-modal representations in medical data.
    \item Extensive experiments show that our framework achieves state-of-the-art results on medical image classification, segmentation, and detection tasks, outperforming existing approaches in medical vision-language pre-training.
\end{itemize}

\section{Related Works}
\subsection{Vision-Language Pre-Training}
Vision-language pre-training (VLP) has gained prominence across various domains by aligning visual and textual representations in a shared feature space, enhancing performance on tasks such as retrieval, classification, and zero-shot learning. Mainstream VLP objectives fall into three categories: contrastive, generative, and matching-based methods. Contrastive objectives \cite{chauhan2020joint, radford2021learning, yang2022unified, zhang2022contrastive} aim to pull paired image-text representations closer in the embedding space while pushing unpaired representations farther apart. Generative objectives \cite{he2022masked, li2023general, wei2022masked}, on the other hand, aim to model cross-modal relationships by generating image or text representations from each other, capturing semantic features through techniques such as image generation, language generation, or cross-modal generation \cite{singh2022flava, zhou2023advancing}. Finally, matching-based methods \cite{dou2022coarse, yao2022detclip, li2022grounded} explicitly match visual and textual elements, either globally (image-text matching) or locally (region-word alignment). Thus, this method requires the ground truth between a word and its corresponding region, to effectively  anchor visual features to textual elements. However, these ground truth labels are often scarce and time-consuming to annotate.

While these methods have shown success with natural images, they often encounter challenges when applied to medical data. Medical images contain subtle, domain-specific features, and clinical reports use complex language, making it difficult to capture nuanced cross-modal relationships with standard VLP approaches. Our contrastive-based method addresses these needs with language-guided, multi-level alignment (i.e. image-report and patch-word), tailored to capture the nuanced relationships in medical images and reports. 

\subsection{Medical Image and Report Pre-Training}
In recent years, pre-training on large-scale medical image-report datasets has become a promising approach to enhancing performance in various medical tasks, including disease classification, segmentation, and detection. Models such as BioViL \cite{bannur2023learning}, MedCLIP \cite{wang2022medclip}, and MaCo \cite{huang2024enhancing} use contrastive learning to align medical images with their corresponding reports, creating a robust foundation for transfer learning across diverse medical applications. Other approaches \cite{chen2022align, wu2023medklip, phan2024decomposing, li2024mlip} incorporate expert knowledge into the pre-training phase to better capture clinical relevance and diagnostic cues. However, these methods frequently treat unpaired image-report samples as strict negatives, which can hinder performance, especially for normal cases where medical images and reports from different patients may exhibit high visual and semantic similarity.

To address this, recent methods like MGCA \cite{wang2022multi} and MLIP \cite{li2024mlip} leverage the Sinkhorn-Knopp clustering algorithm \cite{cuturi2013sinkhorn} to generate cluster assignments for inter-subject semantic correspondences, but this approach does not fully resolve the limitations posed by false negatives in high-similarity cases. Drawing inspiration from multi-granularity frameworks, we propose a Semantics-aware Instance-level Alignment (SIA) module to address the false negative issue by incorporating inter-report relationships into the pre-training phase. Additionally, we introduce a Sparse Token-level Alignment (STA) module to capture fine-grained patch-level correspondences between images and reports, enabling a deeper alignment that reflects specific diagnostic features. Together, these modules improve representation quality and enhance performance in fine-grained medical tasks like segmentation and classification, effectively bridging gaps in existing VLP methods for medical data.

\section{Method}

\begin{figure*}[!ht]
  \centering
  \includegraphics[width=\textwidth]{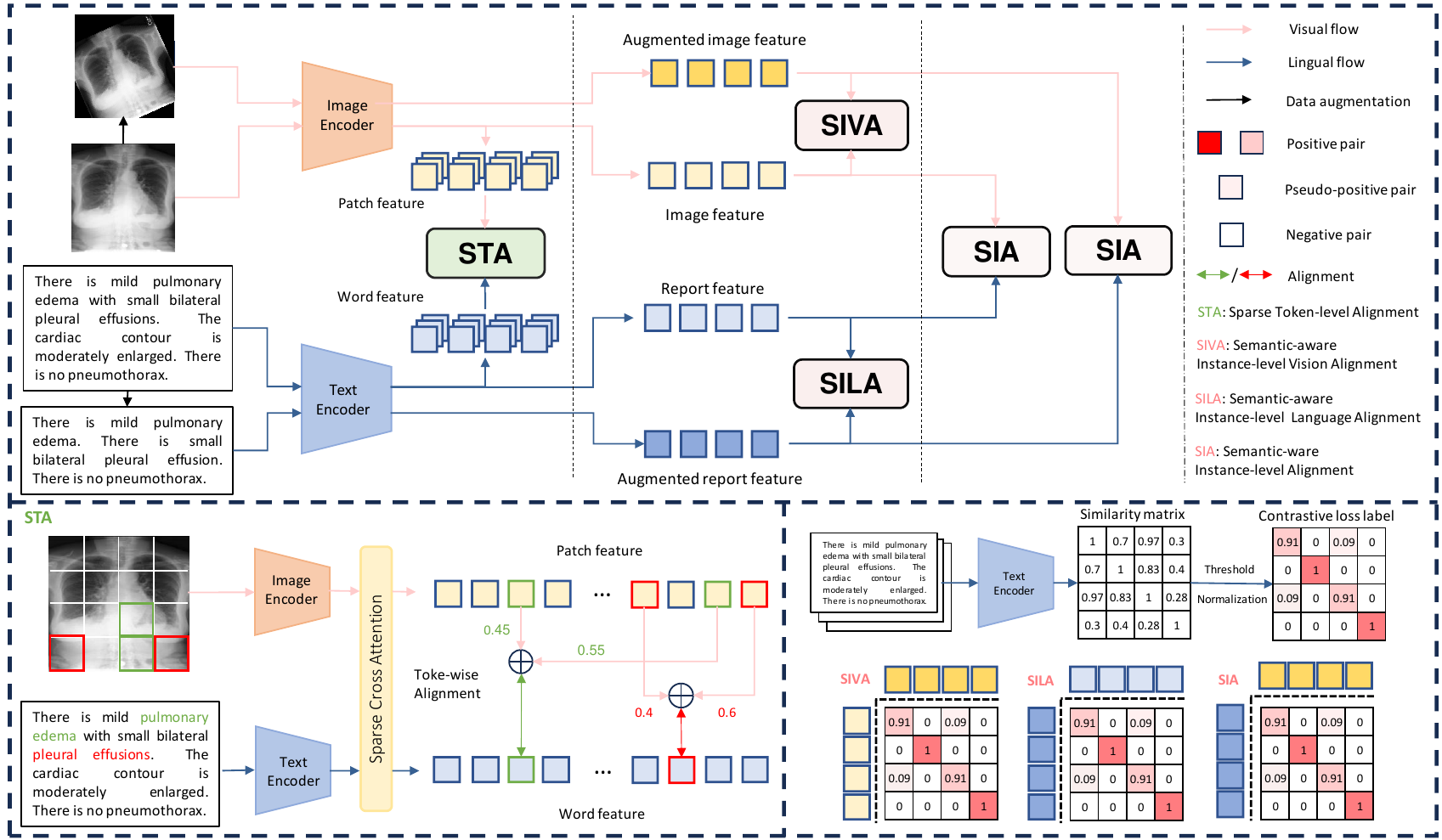}
    \caption{Overview of our proposed SISTA framework for medical image-report alignment. The framework includes an image encoder and a text encoder to capture multimodal representations from chest X-rays and their corresponding reports. Sparse Token Alignment (STA) aligns specific regions in the image with disease tokens via sparse cross-attention, creating token-level correspondence. Semantic-aware Instance-level Vision Alignment (SIVA) and Semantic-aware Instance-level Language Alignment (SILA) ensure consistent instance-level alignment between images and reports.}
    \label{fig: overall_architecture}
\end{figure*}

In this section, we describe the proposed framework, namely SISTA, for medical visual representation learning using chest X-ray images and radiology reports, operating at both instance and token levels. As illustrated in Figure. \ref{fig: overall_architecture}, our SISTA framework consists of five main components: (1-2) semantics-aware instance-level alignment (SIA) modules for original and augmented data, (3) semantics-aware instance-level vision alignment (SIVA) module, (4) semantics-aware instance-level language alignment (SILA) module, and (5) sparse token-level alignment (STA) module. In the following subsections, we provide a detailed description of each component and the overall training objective.

\subsection{Preliminaries}
We aim to to learn robust and generalized medical image representations from radiology reports based on contrastive learning to enhance the performance in downstream tasks such as medical image classification, segmentation, and object detection. Given a training set of \textit{N} image-report pairs $\mathcal{D} = \{(I_1, R_1), (I_2, R_2), ... (I_N, R_N)\}$, we employ an image encoder $f_i$ and a text encoder $f_t$ to extract the features at multiple levels. Following \cite{chen2020simple}, for the i-th image, two non-linear projection layers are employed to transform the image encoder's output into normalized lower-dimensional embeddings, resulting in global and local image features $(v_i, p_{i,1}, p_{i,2}, ..., p_{i,M})$ and the augmented global features $\hat{v}_i$. Similarly, we use a text encoder $f_t$ and two non-linear projection layers to obtain the global and local report features $(t_i, w_{i,1}, w_{i,2}, ..., w_{i,L})$ and the augmented global report feature $\hat{t}_i$. Here, $M$ and $L$ represent the number of patches in the image and the token length, respectively. Finally, we define the cosine similarity function used to compute the contrastive loss between two features \(u\) and \(v\) as follows: 
\[
sim(u, v)= \phi\left(u, v\right) = \frac{\overline{u}}{\left\|\overline{u}\right\|_2} \cdot \frac{\overline{v}}{\left\|\overline{v}\right\|_2}
\]

\subsection{Semantics-aware Instance-level Alignment}
Conventional contrastive learning methods \cite{wang2022multi, li2024mlip} aim to bring paired samples closer and push unpaired samples further apart in the latent space and ignores the similarity between them, which may lead to suboptimal representation learning. To overcome this issue, the proposed SIA module aims to address the false negatives problem by utilizing the radiology report to recalculate the similarity matrix between images in a batch. Considering a target image-report pair, we calculate the textual similarity scores between the target report and others in a batch using the global report features. We employ a threshold (which was set to 0.9) to adopt new pseudo-positives for the target pair. Unlike the original similarity matrix, where the similarity scores for the target pair were set to 0 (hard label), we set the similarity score for pseudo-positives to 0.1 (soft label). This matrix, denoted as \(S\), is then normalized and adopted for the pre-training process. This process is optimized with an image-text contrastive loss based on the InfoNCE loss \cite{oord2018representation}, designed to maximize mutual information between ``positive" image-text pairs in the latent space, as follows:
\[
\begin{aligned}
& \mathcal{L}_{nce}\left(v_i, t_i\right)=-\log \frac{u_{i,i}\exp(\phi \left(v_i, t_i\right) / \tau)}{\sum_{k=1}^B u_{i,k}\exp(\phi\left(v_i, t_k\right)/ \tau)} \\
& \mathcal{L}_{nce}\left(t_i, v_i\right)=-\log \frac{u_{i,i}\exp(\phi \left(t_i, v_i\right) / \tau)}{\sum_{k=1}^B u_{i,k}\exp(\phi\left(v_k, t_i\right)/ \tau)}
\end{aligned}
\]
where $u_{i,k}$ is the language-guided score in the similarity matrix \(S\), $B$ is the batch size and $\tau$ is the global temperature hyper-parameters. The overall objective of the SIA module is the average of the two losses:
\[
\mathcal{L}_{SIA} = \frac{1}{2N} \sum_{i=1}^N(\mathcal{L}_{nce}\left(v_i, t_i\right) + \mathcal{L}_{nce}\left(t_i, v_i\right))
\]

We also extend the SIA module to operate on augmented versions of both images and radiology reports. In the vision branch, various data augmentations are applied, including random rotation, color jitter, and Gaussian blur. For the language branch, we leverage large language models to generate concise summaries of the reports, creating semantically rich augmented versions. The contrastive loss \(\mathcal{L}_{SIA^a}\) is then calculated similarly, as follows:
\[
\mathcal{L}_{SIA^a} = \frac{1}{2N} \sum_{i=1}^N(\mathcal{L}_{nce}(\hat{v}_i, \hat{t}_i) + \mathcal{L}_{nce}(\hat{t}_i, \hat{v}_i))
\]

\subsection{Intra-modal Semantics-aware Instance-level Alignment}
Building upon self-supervised learning approaches in the vision domain \cite{chen2020simple, he2020momentum}, we introduce two intra-modal instance-level alignment modules, tailored separately for the vision and language branches, as follows:

\textbf{Semantics-aware Instance-level Vision Alignment (SIVA):} In the vision branch, we utilize the similarity matrix \(S\), computed by the SIA module, to enhance contrastive learning between original and augmented image representations. Both images are encoded into a shared space, producing global features \(v_i\) and \(\hat{v}_i\), respectively. The contrastive loss in the vision branch, \(\mathcal{L}_{SIVA}\), is computed between \(v_i\) and \(\hat{v}_i\) as shown in the equation below. The language-guided contrastive learning leads the model to treat semantically similar images as “soft negatives” rather than strict negatives. By incorporating both original and augmented views, the model learns to maintain consistency across representations, leading to refined visual embeddings that improve performance on downstream tasks such as image classification, image segmentation, and object detection.
\[
\mathcal{L}_{SIVA} = \frac{1}{2N} \sum_{i=1}^N(\mathcal{L}_{nce}\left(v_i, \hat{v}_i\right) + \mathcal{L}_{nce}\left(\hat{v}_i, v_i\right))
\]

\textbf{Semantics-aware Instance-level Language Alignment (SILA):} Similarly, we apply the similarity matrix \(S\), computed by the SIA module, to enhance the contrastive learning of original and augmented radiology report representations. For each report \(R_i\), an augmented version \(\hat{R}_i\) is generated using large language models, which preserve the semantic integrity of the report. The contrastive loss in the language branch, \(\mathcal{L}_{SILA}\), is computed between each original report feature \(t_i\) and its augmented counterpart \(\hat{t}_i\) as shown in the equation below. This approach of combining original and augmented reports encourages the model to maintain consistent textual representations across different syntactic variations, capturing the stable semantic core of each report. By integrating the similarity matrix to identify and mitigate false negatives, the language branch achieves a more meaningful and robust alignment within the textual domain, enabling the model to better capture nuanced relationships between medical terms and conditions, and ultimately improving its transferability for language-based downstream tasks.
\[
\mathcal{L}_{SILA} = \frac{1}{2N} \sum_{i=1}^N(\mathcal{L}_{nce}(t_i, \hat{t}_i) + \mathcal{L}_{nce}(\hat{t}_i, t_i))
\]

\subsection{Sparse Token-level Alignment}
Considering that pathologies in medical images are often subtle and localized, occupying only a small portion of the image, and that only a few specific terms in the report accurately describe these critical findings, we introduce the Sparse Token-level Alignment (STA) module to complement instance-level alignment. The STA module operates at a finer granularity, aligning individual image patches with specific word tokens from radiology reports to establish precise patch-word correspondences. This alignment approach facilitates fine-grained feature learning by ensuring that critical visual features are effectively aligned with relevant textual descriptions. Unlike prior work \cite{wang2022multi, li2024mlip, mukhoti2023open}, which builds localized visual representations using attention-weighted patch embeddings with text tokens, the STA module directly learns sparse and selective patch-word alignments, bypassing softmax-based attention weighting method.

In the proposed STA module, we select the most relevant patches for alignment with each text token through sparse cross attention mechanism. We begin by dividing each medical image \(I_i\) into a set of patches \(\{p_{i,k}\}_{k=1}^{M}\), where \(M\) is the total number of patches. Simultaneously, each radiology report \(R_i\) is tokenized into words \(\{w_{i,l}\}_{l=1}^{L}\), where \(L\) is the length of the report in tokens. First, we compute the similarity set $\mathbf{S}^i_l = \{s^i_{l,k}\}_{k=1}^M$ between each text token \(w_{i,l}\) and all image patches \(\{p_k\}_{k=1}^{M}\) using inner product (\(s^i_{l,k} = w_{i,l} \cdot p_{i,k}^T\)). To obtain alignment weights, these similarities are normalized to the range \([0, 1]\) using min-max normalization as follows:
\[
\hat{s}_{l,k}^i = \frac{s^i_{l,k} - \min \mathbf{S}^i_l}{\max \mathbf{S}^i_l - \min \mathbf{S}^i_l}
\]

Next, patches with normalized similarities above an empirically determined threshold (e.g. 0.3) are retained, promoting a sparse alignment that reduces computational complexity and focuses on critical image regions. For each retained patch, its final alignment weight $a^i_{l,k}$ is calculated by dividing its similarity score by the total sum of all $R$ retained similarities for that token as follows:
\[
a_{l,k}^i = \frac{\hat{s}^i_{l,k}}{\sum_{r=1}^R \hat{s}^i_{l,r}}
\]
These alignment weights are then used to formulated the cross-modal word embeddings \(\hat{w}_l\) as follows:
\[
\hat{w}_{i,l} = \sum_{r=1}^R a^i_{l,r} * p_{i,r}
\]

Similar to SIA module, we leverage InfoNCE loss \cite{oord2018representation} in the STA module by applying token-level contrastive loss to maximize the mutual information between each word token \(w_{i,l}\) and its weighted combination of the selected image patches \(\hat{w}_{i,l}\) and push unrelated word embeddings far apart. Recognizing that different word tokens vary in importance \cite{wang2022multi, li2024mlip} (e.g., tokens describing pathologies are significantly more relevant than those containing irrelevant information), we assign an importance weight \(u_{i,l}\) to each \(l\)-th token in the report \(R_i\). Thus, the STA loss is formulated as follows:
\[
\mathcal{L}_{nce}^i\left(w_{i,l}, \hat{w}_{i,l}\right) = \frac{1}{L_i} \sum_{l=1}^{L_i}u_{i,l}\log \frac{\exp (\frac{\phi(w_{i,l}, \hat{w}_{i,l})}{\tau})}{\sum_{k=1}^{L_i} \exp (\frac{\phi(w_{i,l}, \hat{w}_{i,k})}{\tau})}\]
\[
\mathcal{L}_{nce}^i\left(\hat{w}_{i,l}, w_{i,l}\right) = \frac{1}{L_i} \sum_{l=1}^{L_i}u_{i,l}\log \frac{\exp (\frac{\phi(\hat{w}_{i,l}, w_{i,l})}{\tau})}{\sum_{k=1}^{L_i} \exp (\frac{\phi(\hat{w}_{i,k}, w_{i,l})}{\tau})}
\]
\[
\mathcal{L}_{STA} = -\frac{1}{2 N}\sum_{i=1}^N \left(\mathcal{L}_{nce}^i\left(w_{i,l}, \hat{w}_{i,l}\right) + \mathcal{L}_{nce}^i\left(\hat{w}_{i,l}, w_{i,l}\right)\right)
\]

The combination of sparse grouping with token-level contrastive learning allows the STA module to capture specific patch-word associations, focusing on the most semantically relevant parts of the image. This sparse alignment strategy improves efficiency and enhances the quality of learned representations by selectively attending to critical image regions for each term in the report.

\subsection{Overall objective function}
Our training approach involves joint optimization of the five losses, designed to encourage the network to learn robust and generalizable representations of medical images. Optimizing the overall loss biases the learned representation to retain fine-grained details associated with each image, as captured in the radiology report, rather than relying solely on global features sufficient for minimizing the global contrastive loss. The overall training objective can be expressed as follows:
\[
\begin{aligned}
\mathcal{L}_{total} = \mathcal{L}_{SIA} + \mathcal{L}_{SIA^a} + \mathcal{L}_{SIVA} + \mathcal{L}_{SILA} + \mathcal{L}_{STA}
\end{aligned}
\]

\section{Experiments}

\subsection{Pre-training setup}
We pre-train the SISTA framework on the MIMIC-CXR 2.0.0 dataset \cite{johnson2019mimic}, applying preprocessing methods from \cite{wang2022multi, li2024mlip} to ensure data consistency. Lateral views are excluded to maintain alignment with downstream datasets, which contain only frontal-view chest images. Following \cite{zhang2022contrastive}, the impression and findings sections were extracted from radiology reports to capture comprehensive descriptions of medical conditions. To ensure data quality, reports that are either empty or too brief are excluded, yielding a dataset of approximately 217,000 image-text pairs. Following \cite{wang2022multi, li2024mlip}, we employ BioClinicalBERT \cite{alsentzer2019publicly} and ResNet-50 \cite{he2016deep} or ViT-B/16 \cite{dosovitskiy2020image} as the text and image encoders, respectively. The framework is trained for 50 epochs on two RTX A100 GPUs with a batch size of 256, using early stopping if the total validation loss does not improve for 5 epochs. We optimize using AdamW \cite{loshchilov2017fixing}, with a learning rate of $2e-5$ and weight decay of 0.05. A linear warm-up followed by a cosine annealing scheduler \cite{loshchilov2016sgdr} is applied, initializing the learning rate at $1e-8$ and setting the warm-up period to 10 epochs. We set the feature dimension $d$ to 128 and the temperature hyperparameters \(\tau\) to 0.2. 

\subsection{Downstream setup}
\textbf{Medical image classification:} We perform classification experiments on three medical datasets: (1) CheXpert \cite{irvin2019chexpert}, consisting of 191,229 frontal chest radiographs, where each image is labeled for five conditions: pleural effusion, edema, atelectasis, cardiomegaly and consolidation. Consistent with prior studies \cite{zhang2022contrastive, wang2022multi, li2024mlip}, we utilize the validation set as the test set and randomly select 5,000 images from the training set for validation. (2) RSNA Pneumonia \cite{shih2019augmenting}, comprising  ~29,700 frontal-view chest X-rays, is used for binary classification between normal and pneumothorax-positive cases. This dataset is partitioned into training, validation, and test sets in a 70\%/15\%/15\% ratio. (3) COVIDx \cite{wang2020covid}, containing over 30,000 chest X-rays, designed for three-class classification into non-COVID pneumonia, COVID-19, and normal categories. We retain the original validation set for testing and allocate 10\% of the training set as a validation subset. For a fair comparison we strictly follow the training protocols in \cite{wang2022multi, li2024mlip}. For fine-tuning, we freeze the image encoder (ResNet-50 or ViT-B/16) and train a linear classification layer on top. The performance is evaluated using AUC score for the CheXpert and RSNA datasets, and accuracy (ACC) for COVIDx.

\begin{table*}[!ht]
  \centering
  \setlength{\tabcolsep}{8pt}
  \caption{Linear classification results on the CheXpert, RSNA, and COVIDx datasets, with fine-tuning conducted on 1\%, 10\%, and 100\% of the training data. Symbols $\dag$ and $\ddag$ indicate methods that incorporate instance-level labels and auxiliary knowledge, respectively. The best and second-best results are highlighted in \textbf{bold} and \underline{underline}, respectively.}
  \resizebox{\linewidth}{!}{
  \begin{tabular}{lcccccccccc}
    \toprule
    \multirow{2}{*}{Method} & \multirow{2}{*}{Venue/Year} & \multicolumn{3}{c}{CheXpert (AUC)} & \multicolumn{3}{c}{RSNA (AUC)} & \multicolumn{3}{c}{COVIDx (ACC)} \\
           & & 1\% & 10\% & 100\% & 1\% & 10\% & 100\% & 1\% & 10\% & 100\% \\
    \midrule
    Random & - & 56.1 & 62.6 & 65.7 & 58.9 & 69.4 & 74.1 & 50.5 & 60.3 & 70.0 \\
    ImageNet & - & 74.4 & 79.7 & 81.4 & 74.9 & 74.5 & 76.3 & 64.8 & 78.8 & 86.3 \\
    GLoRIA (RN-50) \cite{huang2021gloria} & ICCV'21 & 87.1 & 88.0 & 88.2 & 87.0 & \underline{89.4} & 90.2 & 67.3 & 81.5 & 88.6 \\ 
    ConVIRT (RN-50) \cite{zhang2022contrastive} & PMLR'22 & 85.9 & 86.8 & 87.3 & 77.4 & 80.1 & 81.3 & 72.5 & 82.5 & \textbf{92.0} \\ 
    MGCA (RN-50) \cite{wang2022multi} & NIPS'22 & 87.6 & 88.0 & 88.2 & 88.6 & 89.1 & 89.9 & 72.0 & 83.5 & 90.5 \\ 
    M-FLAG (RN-50) \cite{liu2023m} & MICCAI'23 & \underline{87.8} & 88.4 & 88.6 & \underline{88.8} & \underline{89.4} & \underline{90.5} & 72.2 & 84.1 & 90.7 \\ 
    PRIOR (RN-50) \cite{cheng2023prior} & ICCV'23 & 86.2 & 88.3 & 88.6 & 85.7 & 87.1 & 89.2 & 72.3 & 84.7 & \underline{91.0} \\ 
    $\text{MedKLIP}^{\dag}$ (RN-50) \cite{wu2023medklip} & ICCV'23 & 86.2 & 86.5 & 87.7 & 87.3 & 88.0 & 89.3 & \textbf{74.5} & \underline{85.2} & 90.3 \\ 
    $\text{MLIP}^{\ddag}$ (RN-50) \cite{li2024mlip} & CVPR'24 & \underline{87.8} & \textbf{88.7} & \textbf{88.9} & \underline{88.8} & \textbf{89.6} & \textbf{90.6} & \underline{73.0} & 85.0 & 90.5 \\ 
    \rowcolor{Gray}\textbf{SISTA} (RN-50) & - & \textbf{88.1} & \underline{88.6} & \underline{88.7} & \textbf{88.9} & \textbf{89.6} & \textbf{90.6} & 72.3 & \textbf{86.0} & \textbf{92.0}\\
    \midrule
    MGCA (ViT-B/16) \cite{wang2022multi} & NIPS'22 & \underline{88.8} & 89.1 & \underline{89.7} & \underline{89.1} & \underline{89.9} & \textbf{90.8} & 74.8 & 84.8 & 92.3 \\ 
    $\text{MLIP}^{\ddag}$ (ViT-B/16) \cite{li2024mlip} & CVPR'24 & \textbf{89.0} & \textbf{89.4} & \textbf{90.0} & \textbf{89.3} & \textbf{90.0} & \textbf{90.8} & \textbf{75.3} & \underline{86.3} & \underline{92.5} \\ 
    \rowcolor{Gray}\textbf{SISTA} (ViT-B/16) & - & 88.7 & \underline{89.2} & \textbf{90.0} & \textbf{89.3} & \underline{89.9} & \underline{90.6} & \underline{75.0} & \textbf{86.5} & \textbf{93.0}\\
    \bottomrule
  \end{tabular}
  }
  \label{tab:1}
\end{table*}

\textbf{Medical image segmentation:} Our segmentation evaluation is performed on the SIIM Pneumothorax \cite{zawacki2019siim} and RSNA Pneumonia datasets. (1) SIIM Pneumothorax comprises 12,047 chest radiographs with with pneumothorax segmentation masks, divided into training (70\%), validation (15\%), and test (15\%) sets. (2) RSNA Pneumonia \cite{shih2019augmenting} consists of 29,700 chest X-rays, which is partitioned into training (16,010 images), validation (5,337 images), and test (5,337 images) sets. We employ a U-Net with a pre-trained frozen backbone and fine-tune only the decoder \cite{ronneberger2015u}. For a fair comparison, we adhere the training settings in \cite{wang2022multi, li2024mlip}. Dice score serves as the metric for evaluating performance.

\textbf{Medical object detection:} We assess object detection performance on two datasets: (1) RSNA Pneumonia \cite{shih2019augmenting}, comprising 29,700 chest X-rays with bounding box annotations for pneumonia, following the standard data split. (2) Object CXR, a dataset of 9,000 frontal-view chest X-rays annotated with bounding boxes for foreign objects, divided into 6,400 training, 1,600 validation, and 1,000 test images. We adopt a YOLOv3 architecture \cite{farhadi2018yolov3} with a pre-trained ResNet50 frozen backbone, fine-tuning only the detection layers. We evaluate model performance using Mean Average Precision (mAP), calculated over Intersection over Union (IoU) thresholds ranging from 0.4 to 0.75.

\subsection{Performance evaluation}

\begin{table}[!ht]
  \centering
  \caption{Image segmentation results on the SIIM and RSNA datasets, with fine-tuning conducted on 1\%, 10\%, and 100\% of the training data. Symbols $\dag$ and $\ddag$ indicate methods that incorporate instance-level labels and auxiliary knowledge.}
  \resizebox{\linewidth}{!}{
  \begin{tabular}{lcccccc}
    \toprule
    \multirow{2}{*}{Method}  & \multicolumn{3}{c}{SIIM (Dice)} & \multicolumn{3}{c}{RSNA (Dice)} \\
        & 1\% & 10\% & 100\% & 1\% & 10\% & 100\% \\
    \midrule
    Random & 9.0 & 28.6	& 56.3 & 6.9 & 10.6 & 18.5 \\
    ImageNet & 10.2 & 35.5 & 63.5 & 34.8 & 39.9 & 64.0 \\
    ConVIRT (RN-50) & 25.0 & 43.2 & 59.9 & 55.0 & 67.4 & 67.5 \\ 
    GLoRIA (RN-50)  & 37.6 & 56.4 & 64.0 & 60.8 & 68.2 & 67.6 \\ 
    MGCA (RN-50) & 49.7 & 59.3 & 64.2 & 63.0 & 68.3 & 69.8 \\ 
    M-FLAG (RN-50) & \underline{52.5} & \underline{61.2} & 64.8 & 64.6 & \underline{69.7} & 70.5 \\ 
    PRIOR (RN-50) & 51.2 & 59.7 & 66.3 & 66.4 & 68.3 & 72.7 \\ 
    MedKLIP $\dag$ (RN-50) & 50.2 & 60.8 & 63.9 & 66.2 & 69.4 & 71.9 \\ 
    MLIP $\ddag$ (RN-50) & 51.6 & 60.8 & \underline{68.1} & \textbf{67.7} & 68.8 & \underline{73.5} \\ 
    \rowcolor{Gray}\textbf{SISTA} (RN-50) & \textbf{64.6} & \textbf{62.9} & \textbf{72.1} & \underline{67.2} & \textbf{70.6} & \textbf{74.5} \\
    \midrule
    MGCA (ViT-B/16) & 49.0 & 64.7 & 66.4 & 66.2 & 71.3 & 73.6 \\ 
    \rowcolor{Gray}\textbf{SISTA} (ViT-B/16) & \textbf{61.5} & \textbf{68.1} & \textbf{68.7} & \textbf{68.7} & \textbf{73.0} & \textbf{75.1} \\
    \bottomrule
  \end{tabular}
  }
  \label{tab:2}
\end{table}

\textbf{Medical image classification:} As shown in Table \ref{tab:1}, SISTA achieves an AUC of 88.1\% in the 1\% data setting, outperforming MLIP \cite{li2024mlip} by 0.4\% on the CheXpert dataset. In terms of the RSNA dataset, we achieve an AUC score of 88.9\%, 89.6\% and 90\% in the 1\%, 10\% and 100\% data settings, respectively. This marks an improvement of 0.1\% of the previous best-performing model, MLIP, in the 1\% data scenario. Unlike MLIP \cite{li2024mlip}, which leverages external domain knowledge from UMLS, SISTA attains competitive results without relying on additional external resources. On the COVIDx dataset, our model attains a 86.0\% and 92\% accuracy in the 10\% and 100\% data settings, surpassing MedKLIP \cite{wu2023medklip} by 0.8\% and 1.7\%, respectively. Our method also achieves similar or better performance on ViT-B/16 backbone. Please note that the results of these approaches are obtained from original papers.

\textbf{Medical image segmentation:} Table \ref{tab:2} shows the segmentation results on the SIIM and RSNA datasets across different data settings (1\%, 10\% and 100\% of training  data). On the SIIM dataset, the proposed SISTA outperforms other state-of-the-art models, M-FLAG \cite{liu2023m} and MLIP \cite{li2024mlip}, 12.1\%, 1.7\%, 4.0\% in the 1\%, 10\% and 100\% settings, respectively. On the RSNA dataset, SISTA surpasses the previous top model, MLIP, by 1.8\% in the 10\% setting and 1\% in the 100\% setting with ResNet-50 as the backbone. Our method also performs better than MGCA method on ViT-B/16 backbone. These results indicate  that SISTA consistently achieves state-of-the-art performance across multiple data settings, and demonstrates strong generalizability across different backboens on both segmentation datasets.

\begin{table}[!ht]
  \centering
  \caption{Object detection results on the RSNA and Object CXR datasets, fine-tune on 1\%, 10\%, and 100\% of the training data. $\dag$ and $\ddag$ indicate methods that incorporate instance-level labels and auxiliary knowledge, respectively. ``$\thicksim$" indicates mAP $<$ 1\%.}
  \resizebox{\linewidth}{!}{
  \begin{tabular}{lcccccc}
    \toprule
    \multirow{2}{*}{Method} & \multicolumn{3}{c}{RSNA (mAP)} & \multicolumn{3}{c}{Object CXR (mAP)} \\
        & 1\% & 10\% & 100\% & 1\% & 10\% & 100\% \\
    \midrule
    Random & 1.0 & 4.0 & 8.9 & $\thicksim$ & $\thicksim$ & 4.4 \\
    ImageNet & 3.6 & 8.0 & 15.7 & $\thicksim$ & 8.6 & 15.9 \\
    ConVIRT & 8.2 & 15.6 & 17.9 & $\thicksim$ & 8.6 & 15.9 \\ 
    GLoRIA & 10.3 & 15.6 & 23.1 & $\thicksim$ & 8.9 & 16.6 \\ 
    MGCA & 12.9 & 16.8 & 24.9 & $\thicksim$ & 12.1 & 19.2 \\ 
    M-FLAG & 13.7 & 17.5 & 25.4 & $\thicksim$ & 13.6 & 19.5 \\ 
    PRIOR & 15.6 & 18.5 & 25.2 & 2.9 & 15.2 & 19.8 \\ 
    MedKLIP $\dag$ & 8.9 & 16.3 & 24.5 & $\thicksim$ & 7.1 & 11.6 \\ 
    MLIP $\ddag$ & \underline{17.2} & \underline{19.1} & \underline{25.8} & \underline{4.6} & \underline{17.4} & \underline{20.2} \\ 
    \rowcolor{Gray} \textbf{SISTA} & \textbf{22.0} & \textbf{24.3} & \textbf{27.4} & \textbf{11.0} & \textbf{22.9} & \textbf{20.9} \\
    \bottomrule
  \end{tabular}
  }
  \label{tab:3}
\end{table}

\textbf{Medical object detection:} Table \ref{tab:3} presents the object detection results on the RSNA and Object CXR datasets, evaluated using 1\%, 10\%, and 100\% of the available training data. On the RSNA dataset, SISTA achieves an mAP improvement of 4.8\%, 5.2 \% and 1.6\%, over the previous top-performing model, MLIP, in the 1\%, 10\% and 100\% data settings, respectively. For the Object CXR dataset, SISTA also establishes new benchmarks, surpassing MLIP by 6.4\% in the 1\% setting, 5.5\% in the 10\% setting and 0.7\% in the 100\% setting. These results highlight SISTA's effectiveness and adaptability in object detection tasks, particularly in scenarios with limited labeled data, where it achieves substantial performance gains over competing models. to enhance robust and generalizable medical image representations. to enhance robust and generalizable medical image representations. 

\begin{table*}
  \centering
  \setlength{\tabcolsep}{10pt}
  \caption{Ablation study of our SISTA on three tasks: classification on CheXpert, segmentation on SIIM, and object detection on RSNA dataset. The best and second-best results are highlighted in \textbf{bold} and \underline{underline}, respectively.}
  \begin{tabular}{lcccccccccccccccc}
    \toprule
    \multicolumn{5}{c}{Loss Settings} & \multicolumn{3}{c}{CheXpert (AUC)} & \multicolumn{3}{c}{SIIM (Dice)} & \multicolumn{3}{c}{RSNA (mAP)} \\
    SIA & \(\text{SIA}^a\) & SIVA & SILA & STA & 1\% & 10\% & 100\% & 1\% & 10\% & 100\% & 1\% & 10\% & 100\% \\
    \midrule
    \checkmark & \checkmark & & & & 86.7 & 87.1 & 87.4 & 16.8 & 49.6 & 67.3 & 13.3 & 18.3 & 26.3 \\
    \checkmark & \checkmark & & & \checkmark & 86.6 & 87.7 & 88.2 & \underline{31.4} & 55.3 & \underline{70.4} & \underline{15.0} & 18.8 & \underline{27.0} \\
    \checkmark & \checkmark & \checkmark & \checkmark & & \underline{87.4} & \underline{88.2} & \underline{88.4} & 26.4 & \underline{57.5} & 66.8 & 14.6 & \underline{19.5} & 26.5 \\
    \rowcolor{Gray} \checkmark & \checkmark & \checkmark & \checkmark & \checkmark & \textbf{88.1} & \textbf{88.6} & \textbf{88.7} & \textbf{64.6} & \textbf{62.9} & \textbf{72.1} & \textbf{22.0} & \textbf{24.3} & \textbf{27.4} \\
    \bottomrule
  \end{tabular}
  \label{tab:4}
\end{table*}

\subsection{Ablation study}
\begin{figure}[!h]
  \centering
   \includegraphics[width=\linewidth]{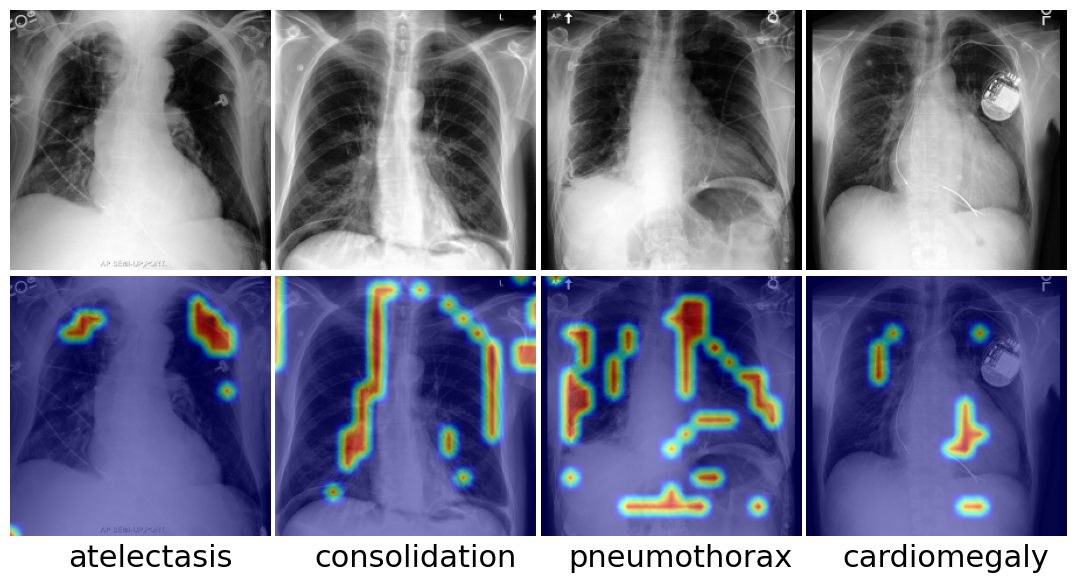}
   \caption{Heatmap visualization of token-level correspondence, with red regions indicating areas of higher activation weights associated with word tokens.}
   \label{fig:heatmap}
\end{figure}

Table \ref{tab:4} summarizes the results of the ablation study on three tasks: medical image classification on the CheXpert dataset, semantic segmentation on the SIIM dataset, and object detection on the RSNA dataset, all using ResNet-50 as the image encoder backbone. Results indicate that the SIVA, SILA, and STA modules improve classification and segmentation performance when combined with the SIA module, suggesting that both intra-modal instance-level and cross-modal token-level alignments enable the image encoder to learn more generalizable representations for downstream tasks. Specifically, the instance-level SIVA and SILA modules primarily boost classification, while the STA module, with its token-level focus, proves particularly effective for segmentation and detection tasks requiring fine-grained information. When all modules are jointly trained, the model achieves optimal performance across all tasks. Figure \ref{fig:heatmap} illustrates our model's ability to capture word-patch and instance-level correspondence in chest X-ray analysis. This figure shows the heatmaps overlaid on chest X-rays, highlighting regions that correspond to specific disease tokens. We visually demonstrate the alignment between textual disease labels and relevant image regions, reflecting token-level correspondence. 

\section{Conclusion}
Medical contrastive vision-language pre-training (VLP) has shown considerable potential in advancing data-efficient learning for downstream tasks. However, existing contrastive methods often treat all unpaired samples as negatives, which can disrupt semantic consistency and weaken representation quality. We introduce SISTA, a multi-level alignment framework for medical VLP that captures semantic correspondences at both image-report and patch-word levels. SISTA integrates two key modules: a Semantics-aware Instance-level Alignment (SIA) module, which mitigates false negatives by leveraging inter-report similarities, and a Sparse Token-level Alignment (STA) module, which aligns specific image patches with corresponding word tokens in radiology reports. The SIA module operates across both cross-modal and intra-modal settings to enhance representation learning, while the STA module establishes sparse groupings of image patches per word token. Experimental results show that SISTA sets a new benchmark in medical VLP, delivering substantial performance gains in image classification, segmentation, and object detection. 

\section*{Acknowledgment}

\bibliographystyle{IEEEtran}
\bibliography{main}

\end{document}